\documentclass[]{fairmeta}

\usepackage[utf8]{inputenc} 
\usepackage[T1]{fontenc}    
\usepackage{hyperref}       
\usepackage{url}            
\usepackage{booktabs}       
\usepackage{amsfonts}       
\usepackage{nicefrac}       
\usepackage{microtype}      
\usepackage{xcolor}         
\usepackage{soul}

\usepackage{subcaption}
\usepackage{graphicx}
\usepackage{tabularx}
\usepackage{geometry}
\usepackage{ragged2e}
\usepackage{array}

\newcolumntype{L}[1]{>{\RaggedRight\hsize=#1\hsize\arraybackslash}X}
\newcolumntype{Y}{>{\Centering\arraybackslash}X}
\usepackage{amsmath} 
\usepackage[most]{tcolorbox}
\usepackage{mdframed}

\newcommand{\ignore}[1]{}

\usepackage[moderate]{savetrees}

\usepackage{caption}
\DeclareCaptionJustification{forcedcenter}{\centering}
\usepackage{pifont}      
\newcommand{\circled}[1]{\ding{\the\numexpr191+#1\relax}}  

\usepackage{titlesec}
\titlespacing*{\paragraph}{0pt}{6pt plus 1pt minus 1pt}{0.5em}

\title{S-EMBER: A Large-Scale Benchmark for Streaming Egocentric Memory Retrieval}

\author[1]{Xiaodong Wang}
\author[1]{Xuanyi Zhao}
\author[1]{Pedro Rodriguez}
\author[1]{Devendra Singh Sachan}
\author[1]{Barlas Oguz}
\author[2]{Seungwhan Moon}
\author[1]{Shang-Wen Li}
\author[1]{Gargi Ghosh}
\author[2]{Xin Dong}
\author[1]{Wen-Tau Yih}

\affiliation[1]{FAIR, Meta}
\affiliation[2]{Reality Labs, Meta}

\abstract{As wearable devices enable continuous first-person recording, AI assistants must reason across long time horizons to recall past experiences—a capability known as episodic memory. Current benchmarks often rely on offline evaluation with access to entire video files, failing to simulate the streaming reality of wearable intelligence. We introduce S-EMBER (Streaming Egocentric Memory Benchmark for Episodic Retrieval), a large-scale benchmark comprising 3,141 videos totaling 388 hours of organic activity captured via Ray-Ban Meta smart glasses. S-EMBER formalizes grounded streaming episodic retrieval, a paradigm shift from global offline search to causal, active recall triggered by visual events in a continuous stream. We provide 9,448 QA pairs requiring manual visual proof through precise temporal localization and supporting flexible response lengths to simulate natural human-AI interaction. Our extensive benchmarking of frontier models reveals a grounded recall gap: models answer and localize with moderate competence in isolation, yet fall furthest short of human performance when both must hold for the same query, the strongest reaching less than half the human rate. S-EMBER establishes a hardware-authentic foundation for developing grounded, reliable episodic memory in the next generation of wearable AI agents.}

\correspondence{Xiaodong Wang at \email{xiaodongwang@meta.com}, Scott Yih at \email{scottyih@meta.com}}

\metadata[Code]{\url{https://github.com/facebookresearch/S-EMBER}}
\metadata[Dataset]{\url{https://huggingface.co/datasets/facebook/S-EMBER}}

\begin{document}

\maketitle

\section{Introduction}
\label{section:intro}
Accurately recalling information about events and reasoning about their temporal relationships---i.e., episodic memory~\citep{tulving1972episodic}---is a crucial capability for personal assistants that aim to process always-on video of everyday events.
For example, suppose we want an agent to answer ``what ingredients do I need to pick up on the way home to bake white chocolate ube cookies again and where can I buy them?''
With the rapid growth of consumer smart glasses\footnote{
 For example, Meta Ray-Bans~\citep{metarayban2023}, Rokid AI \& AR glasses, and XREAL One Pro.
}, assistants could have access to videos of (1) the previous recipe and recording of baking cookies, (2) the contents of ingredients at home to compare to the recipe, and (3) what stores stocked these ingredients last time.
In this example, the assistant must navigate temporal dependencies spread across disparate video segments while filtering for contextually relevant details. To bridge the gap between current offline models and this streaming necessity, we introduce a new egocentric benchmark designed to measure how effectively models locate and utilize information within long-form visual histories (\S\ref{sec:dataset}).

To address these requirements, we introduce S-EMBER (Streaming Egocentric Memory Benchmark for Episodic Retrieval), a large-scale diagnostic foundation comprising 3,141 unique videos totaling 388 hours of organic human activity. Captured via consumer-grade Ray-Ban Meta smart glasses by 613 unique users, S-EMBER provides a hardware-authentic, head-centered perspective that mirrors the actual direction of human attention. 
By moving beyond the fixed, body-mounted rigs characteristic of traditional egocentric datasets, S-EMBER ensures the visual stream aligns with the dynamic, everyday environments that a personal AI assistant must realistically navigate.

Beyond its authentic perspective, S-EMBER fundamentally shifts the evaluation paradigm from the static search found in current VideoQA benchmarks to a causal, streaming recall. We formalize this as grounded streaming episodic retrieval (\S\ref{sec:dataset}), a protocol where 9,448 queries are triggered at specific ``online'' timestamps only when the current visual context creates a logical need for past information. Crucially, each query is paired with a manual visual proof ($[t_{start}, t_{end}]$ interval), enabling the first large-scale verification of whether an agent's reasoning is accurately anchored in the visual history or driven by temporal hallucinations.

Building on this requirement for verification, we organize the benchmark around an eight-category taxonomy of cognitive challenges (Table~\ref{tab:taxonomy}). This framework includes tasks such as Location Trace (e.g., ``Where did I last have my keys?''), Time Duration (e.g., ``How long did the cooking last?''), and Object Comparison (e.g., ``How does this item compare to the one I saw earlier?''). Furthermore, S-EMBER facilitates the study of adaptive memory reporting by providing multi-granular ground-truth responses. This enables a rigorous evaluation of how AI agents should modulate their verbosity---from concise, direct answers to comprehensive structured descriptions---to meet the flexible communication needs of natural human-AI interaction.


Our evaluation of frontier and open-source models reveals a striking grounded recall gap. Models are individually competent at each half of the task: relative to human annotators, the strongest reaches 52\% on answer accuracy and 59\% on interval localization. They are markedly weaker at the conjunction, producing a correct answer drawn from correctly located evidence only 43\% as often as humans, and the best open model only 23\% as often (\S\ref{sec:results}). Increasing model scale, frame density, and resolution improves both halves without closing this gap; grounding in particular saturates early, gaining 3.2 mIoU as the frame budget quadruples from 128 to 512 and 1.3 mIoU as resolution quadruples above 360p (\S\ref{sec:ablation}). These findings indicate that what current architectures lack is neither the ability to understand what happened nor the ability to say when something happened, but the ability to bind the two reliably to the same event.

Beyond this grounding failure, our diagnostic analysis uncovers a universal recall decay as the temporal gap between the evidence and the query increases (\S\ref{sec:diagnostics}). Even high-performing models like Gemini 3.1 Pro experience a steady decline from 50\% accuracy for immediate recall to 29\% for events occurring more than eight minutes prior. This challenge is compounded by an evidence duration bottleneck; while models can effectively retrieve momentary visual ``needles,'' they struggle to aggregate information across extended temporal ranges to form coherent answers. By providing S-EMBER, we offer a rigorous foundation captured from the naturalistic perspective of smart glasses to help the research community move toward reliable, memory-aware personal AI agents.

In summary, we provide a diagnostic, memory-focused foundation for the next generation of wearable intelligence through three primary contributions. First, we introduce S-EMBER, a large-scale benchmark for evaluating streaming episodic retrieval, comprising 3,141 videos (388 hours) and 9,448 QA pairs captured by 613 unique users using consumer-grade smart glasses to mirror natural daily life. Second, we organize this benchmark around a comprehensive eight-category taxonomy that systematically maps diverse episodic memory facets. Every query is paired with a manual interval prediction to serve as objective visual proof, ensuring that model reasoning is authentically anchored in the visual stream rather than driven by temporal hallucinations. 
Finally, we identify a fundamental grounded recall gap across leading model families, including InternVL3.5 and Qwen3VL. Our analysis reveals that models answer and localize with moderate competence in isolation yet succeed at both on the same query at less than half the human rate, and that increases in parameter count, resolution, and frame density improve each capability without closing the distance between them.

\begin{figure}[t]
     \centering
     \includegraphics[width=1.0\textwidth]{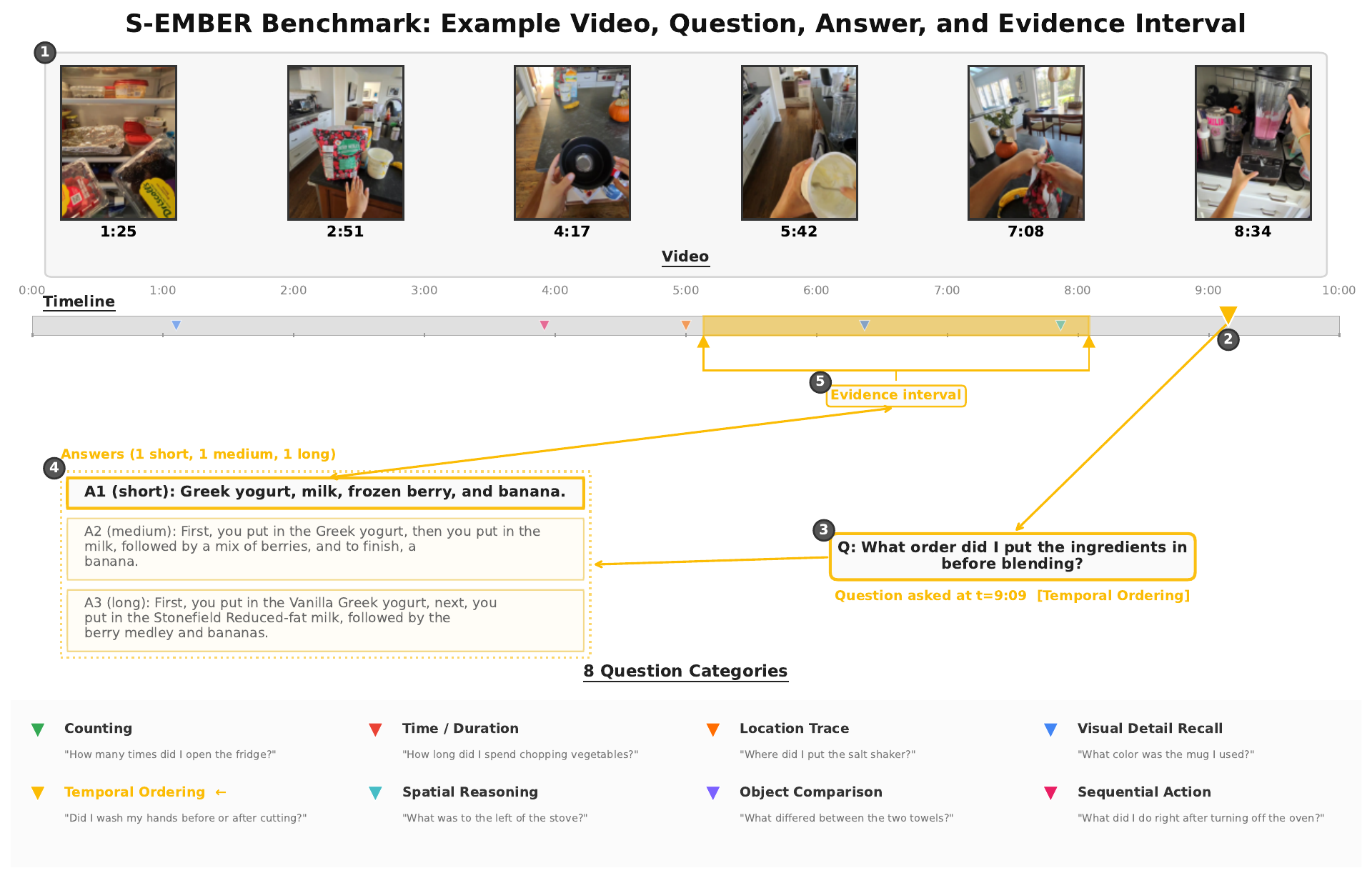}
     \captionsetup{justification=justified}  
     \caption{\textbf{Overview of the S-EMBER benchmark annotation.}
  Annotators issue memory questions in a streaming fashion across 8 color-coded categories. Each question is answered by multiple raters at varying verbosity, and every answer is grounded to a supporting temporal interval that serves as visual memory evidence. Concretely:
  \circled{1} a video, typically 5--20 minutes long;
  \circled{2} a specific point in time at which
  \circled{3} a question is asked (replicating an online/streaming use case);
  \circled{4} three acceptable answers from distinct raters, instructed to write short, medium, and long responses; and
  \circled{5} an evidence interval, marked by the answer writer, indicating the video segment containing the information needed to answer.}
     \label{fig:teaser}
\end{figure}

\section{Related Work}
\label{sec:related_work}

\subsection{Egocentric Video Understanding}
Foundational datasets, particularly Ego4D~\citep{grauman2022ego4d}, have been instrumental in advancing egocentric vision by offering a diverse and unprecedented scale of first-person activity data to the community.
Subsequent benchmarks such as EgoSchema~\citep{mangalam2024egoschema}, EgoTempo~\citep{plizzari2025egotempo}, and AMEGO~\citep{goletto2024amego} have built upon this foundation to test temporal perception and long-range reasoning. As these benchmarks become increasingly well-traversed, there is a burgeoning need for domain shifts that capture higher visual fidelity and new environments. S-EMBER addresses this by providing high-quality footage captured through Ray-Ban Meta smart glasses. Unlike the research-heavy rigs or action cameras used in prior benchmarks like HourVideo~\citep{he2024hourvideo}, X-LeBench~\citep{dai2025xlebench}, or EgoLife~\citep{yang2025egolife}, this form factor represents a shift toward socially compliant, daily-wear hardware, capturing natural gaze and interactions in real-world settings and enabling evaluation of next-generation proactive AI assistants.

\subsection{Long-form Video Reasoning}
Video understanding has undergone a rapid transition from short-clip analysis, exemplified by MVBench~\citep{li2024mvbench} and CinePile~\citep{rawal2024cinepile}, to long-context reasoning. This progress is marked by the emergence of datasets such as Video-MME~\citep{fu2024videomme}, MLVU~\citep{zhou2024mlvu}, LongVideoBench~\citep{wu2024longvideobench}, and InfiniBench~\citep{ataallah2024infinibench}, which push the boundaries of model attention mechanisms by utilizing content and films lasting up to several hours. Other works like LVBench~\citep{wang2024lvbench}, MoVQA~\citep{huang2024movqa}, and Video-MMMU~\citep{hu2025videommmu} have further expanded the diversity of tasks, from multi-hop reasoning to cross-modal knowledge integration. As the field matures, there is an increasing recognition that long-form complexity is defined not solely by temporal duration, but by information density and retrieval difficulty. 
By focusing on unedited life-logging where the ``needle'' of evidence is buried in mundane activities, we shift the challenge from plot-based summarization to precise episodic localization---filling a critical gap between short-form clips and edited long-form content.

\subsection{Grounded Streaming Retrieval}
The ultimate utility of a personal assistant lies in its ability to retrieve information in real time, a paradigm known as streaming retrieval. Unlike traditional offline evaluation where models have access to the entire video, streaming benchmarks like StreamingBench~\citep{li2024streamingbench}, SVBench~\citep{yang2025svbench}, VStream-QA~\citep{huang2024flashvstream}, OVO-Bench~\citep{niu2025ovobench}, and MovieChat-1K~\citep{song2024moviechat} enforce a causal constraint: the model can only see the past and present up to the moment a question is asked. S-EMBER advances this streaming foundation by introducing a more ecological approach to query generation. Rather than using automated scripts or global summaries, our questions are triggered naturally by human annotators as they observe the stream, aligning with the real-world use cases of memory recall.

\begin{table}[t]
\centering
\caption{Comparison of current video benchmarks. Our work represents the largest egocentric dataset using wearable devices with manual annotation and full streaming/interval prediction support. ``Anno.'' indicates the annotation methodology as Manual (M), Automated (A), or hybrid (A/M); ``Task Format'' describes the evaluation types: MC for multiple-choice and OE for open-ended questions; ``Int. Pred.'' denotes whether the benchmark requires temporal interval prediction.}
\label{tab:benchmark-comparison}
\scriptsize 
\setlength{\tabcolsep}{2pt} 
\begin{tabular*}{\textwidth}{@{\extracolsep{\fill}}l ccc c c c l c cc}
\toprule
\textbf{Benchmark} & \textbf{\# Hours} & \textbf{\# Videos} & \textbf{\# Questions} & \textbf{Avg. Len} & \textbf{Anno.} & \textbf{Egocentric} & \textbf{Source} & \textbf{Task Format} & \textbf{Int. Pred.} & \textbf{Stream} \\ \midrule
MVBench & 16.2 & 3,641 & 4,000 & 16s & A/M & No & Multi-source & MC & No & No \\
MovieChat-1K & 20.4 & 130 & 1,950 & 9.4m & M & No & Movies \& TV & MC \& OE & No & Yes \\
VStream-QA & 21.3 & 32 & 3,500 & 40m & M & Partial & Multi-source & MC \& OE & No & Yes \\
MoVQA & 27.5 & 100 & 21,953 & 16.5m & M & No & Movies \& TV & MC & No & No \\
Video-MMMU & 42 & 300 & 900 & 8.4m & M & No & Web & MC & No & No \\
StreamingBench & 61.5 & 900 & 4,500 & 4.1m & M & No & Web (YT) & MC & No & Yes \\
OVO-Bench & 76.2 & 644 & 2,814 & 7.1m & M & No & Web & MC \& OE & No & Yes \\
LVBench & 116.7 & 103 & 1,549 & 68m & M & No & Web (YT) & MC & No & No \\
FALCON-Bench & 121.1 & 92 & 576 & 79m & M & No & Multi-source & MC \& OE & Yes & No \\
Video-MME & 255 & 900 & 2,700 & 17m & M & No & Web (YT) & MC & No & No \\
SVBench & 270.6 & 1,353 & 49,979 & 12m & A/M & No & Web & MC \& OE & No & Yes \\
MLVU & 446.9 & 1,730 & 3,102 & 15.5m & M & No & Web & MC \& OE & No & No \\
LongVideoBench & 489.2 & 3,763 & 6,678 & 7.8m & M & No & Web (YT) & MC & No & No \\
CinePile & 516.8 & 9,396 & 303,828 & 3.3m & A/M & No & Web (YT) & MC & No & No \\
InfiniBench & 1,076.8 & 1,219 & 87,700 & 53m & Auto & No & Movies \& TV & MC & No & No \\ \midrule
EgoTempo & 10 & 798 & 1,000 & 45s & A/M & Yes & Ego4D & MC & No & No \\
AMEGO & 14.2 & 100 & 20,500 & 8.5m & A/M & Yes & Epic-Kitchens & MC & No & No \\
EgoSchema & 253.2 & 5,063 & 5,063 & 3m & A/M & Yes & Ego4D & MC & No & No \\
EgoLife & 265.8 & 6 & 3,000 & 44.3h & A/M & Yes & Aria Glass & MC \& OE & No & No \\
HourVideo & 380.8 & 500 & 12,976 & 45.7m & A/M & Yes & Ego4D & MC & No & No \\
X-LeBench & 288 & 432 & 26,932 & 40m & A/M & Yes & Ego4D & MC & Yes & No \\ \midrule
\textbf{S-EMBER (Ours)} & \textbf{388} & \textbf{3,141} & \textbf{9,448} & \textbf{7.4m} & \textbf{M} & \textbf{Yes} & \textbf{Ray-Ban Meta} & \textbf{MC \& OE} & \textbf{Yes} & \textbf{Yes} \\ \bottomrule
\end{tabular*}
\end{table}

\section{The S-EMBER Dataset}
\label{sec:dataset}

\subsection{Grounded Streaming Episodic Retrieval (GSER)}
We introduce a new evaluation task: Grounded Streaming Episodic Retrieval (GSER).
In this task, assistants receive a video up to the point that a user asks a question.
Like ordinary video question answering (VideoQA) tasks, the assistant should answer correctly, but unlike other tasks, we also require the assistant to identify the time interval $[t_{start}, t_{end}]$ that supports its answer.
To increase the difficulty and test where models fail, we collect questions where $t_{start}$ varies from near the start of the video to near the end; we also instruct raters to write questions that lead to a variety of interval lengths $t_{end}-t_{start}$.

\subsection{Data Collection}
S-EMBER transitions egocentric vision from research rigs to socially compliant, eye-level Ray-Ban Meta smart glasses. This hardware provides a naturalistic, head-centered field of view that mirrors the wearer's actual direction of attention. Our collection strategy involved 613 unique users across diverse real-world settings. By prioritizing unscripted ``in-between'' moments---the mundane intervals of daily life that lack clear narrative beats or high action---we ensure the dataset captures the specific scenarios where episodic memory is most naturally needed.

\subsection{Annotation Pipeline}
After video collection, S-EMBER QA annotations follow a three-stage, causal-first pipeline involving specialized teams for question generation, multi-granular response synthesis, and multi-faceted verification. This design philosophy ensures that every data point comes from a temporal trigger: an event in the video stream ``causes'' a rater to identify a natural point of inquiry, which then drives the downstream answering and validation tasks. Unlike datasets that rely on global video summaries, this workflow enforces a streaming memory constraint, requiring models to navigate long temporal horizons without access to future context or non-visual cues.

\begin{figure}[t]
     \centering
     \includegraphics[width=0.85\textwidth]{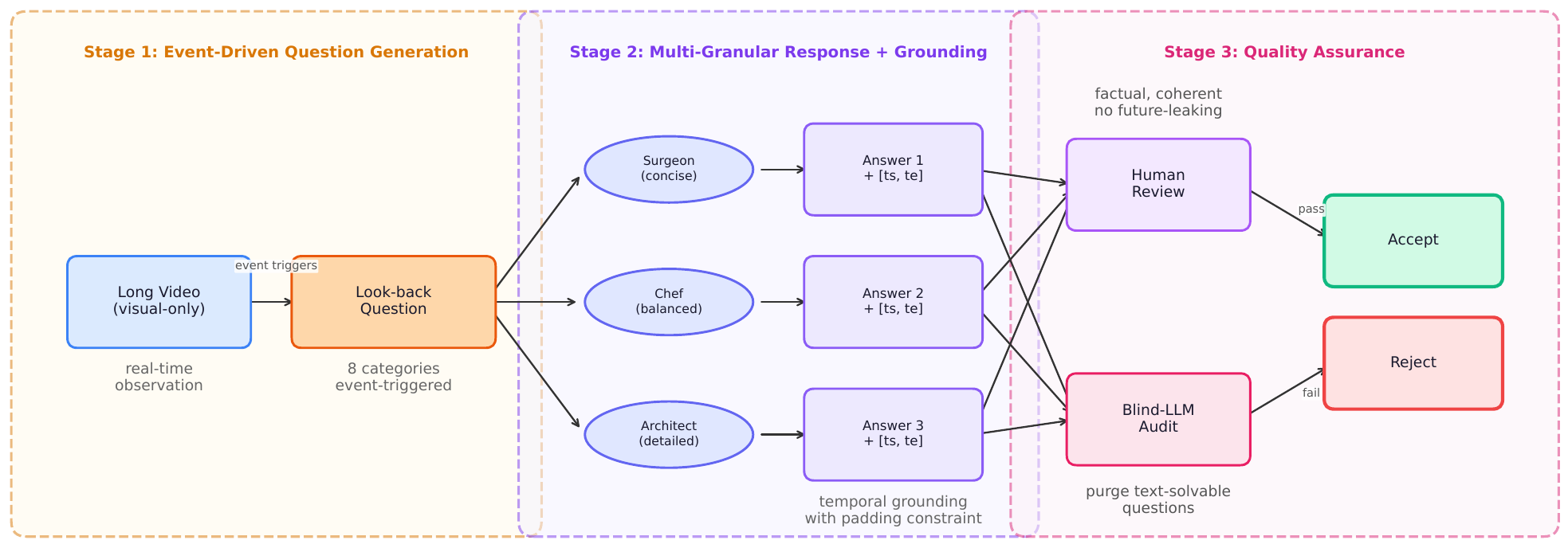}
     \caption{S-EMBER Annotation Workflow.}
     \label{fig:annot}
\end{figure}

\textbf{Stage 1: Generation of Event-Driven Questions.} 
Annotators observe video streams in real time to generate questions that fulfill a ``look-back'' requirement. Every question is triggered by a specific event in the present, but remains unanswerable using only the current visual frame, necessitating retrieval from the past. Raters are instructed to follow the specific prototypes and templates of the eight categories, selecting the most relevant type based on the scene's context. To isolate visual reasoning, the environment is strictly visual-only; audio is muted during both generation and answering. Our filtering protocol further eliminates non-episodic shortcuts, such as questions solvable through external knowledge or predictions of future events.

\textbf{Stage 2: Multi-granular response and temporal grounding.} 
For each validated question, we provide a tri-tiered ground-truth system to facilitate the study of adaptive verbosity. Three independent groups---surgeons (concise), chefs (balanced), and architects (comprehensive)---generate responses that vary in structural complexity while maintaining semantic consensus. Simultaneously, annotators identify the visual proof by marking the precise temporal interval $[t_{start}, t_{end}]$ containing the evidence required to answer the question.
To prevent lazy grounding where an overly broad window might capture the answer by chance, we enforce a strict temporal padding constraint. Any non-essential buffer surrounding the core evidence is restricted to the lesser of 15\% of the evidence's duration or a 15-second maximum. This ensures the ground truth is tightly coupled to the specific visual event rather than relying on broad, safe windows.


\textbf{Stage 3: Independent verification.} The protocol concludes with an independent audit: every question--answer--interval triplet is reviewed by an expert who did not produce it, and is retained only if the question, all three answer tiers, and the supporting interval each pass. Correctness is held to a binary standard, and any triplet failing a criterion is discarded. Agreement before adjudication is high, and is reported in Appendix~\ref{App:protocol}. Finally, we execute a Blind-LLM (o3~\citep{openai2025o3}) audit to identify and purge questions solvable via common-sense priors or linguistic patterns alone, ensuring that success on S-EMBER is contingent on authentic visual episodic grounding rather than dataset shortcuts.

\begin{table}[t]
\caption{S-EMBER Dataset Taxonomy: Cognitive Categories and Question Prototypes}
\centering
\label{tab:taxonomy}
\small
\scalebox{0.9}{ 
\begin{tabularx}{\textwidth}{@{} l L{1.2} L{0.8} @{}}
\toprule
\textbf{Category} & \textbf{Cognitive Focus} & \textbf{Example Question} \\ 
\midrule

Visual Detail Recall & Fine-grained retrieval of attributes or objects visible. & What is the [attribute] of the [object] I saw? \\ \addlinespace[0.6em]

Sequential Action & Understanding the chronological link between two distinct events or states. & After [earlier action], what did I [action] next? \\ \addlinespace[0.6em]

Time Duration & Measuring the continuous or non-consecutive span of a specific activity. & How long did [event/activity] last? \\ \addlinespace[0.6em]

Counting & Quantifying occurrences or objects tracked across multiple temporal windows. & How many times have I done [action] so far? \\ \addlinespace[0.6em]

Temporal Ordering & Step-by-step logical sequencing of complex, multi-stage tasks. & What steps did I take to complete [task]? \\ \addlinespace[0.6em]

Object Comparison & Contrasting attributes (price, size, etc.) between current and historical observations. & How does this [object] compare to the similar one I saw before? \\ \addlinespace[0.6em]

Location Trace & Indexing the history of an object's spatial position or its last known placement. & Where did I last have the [object] before [event]? \\ \addlinespace[0.6em]

Spatial Reasoning & Navigation and mapping of previously visited environments or scene layouts. & How do I get back to [X] from where I am now? \\ 

\bottomrule
\end{tabularx}
}
\end{table}

\subsection{Data Statistics}
S-EMBER comprises 3,141 unique videos averaging 7.4 minutes, resulting in approximately 388 hours of egocentric footage. Videos range from roughly 5 to 20 minutes, reflecting the natural duration of everyday activities. The dataset distribution is purposefully sparse, mirroring natural human inquiries with an average of 3 questions per video, ranging from 2 to 6 questions per session.

The videos span 2,090 unique fine-grained scenario labels (e.g., ``grocery shopping trip'', ``pancake breakfast preparation''), which are consolidated into 33 broad activity categories. 
This distribution reflects the natural frequency of daily activities in egocentric wearable recordings.

\begin{figure}[tbp]
  \centering
  \begin{subfigure}[b]{0.49\textwidth}
      \centering
      \includegraphics[width=0.9\textwidth]{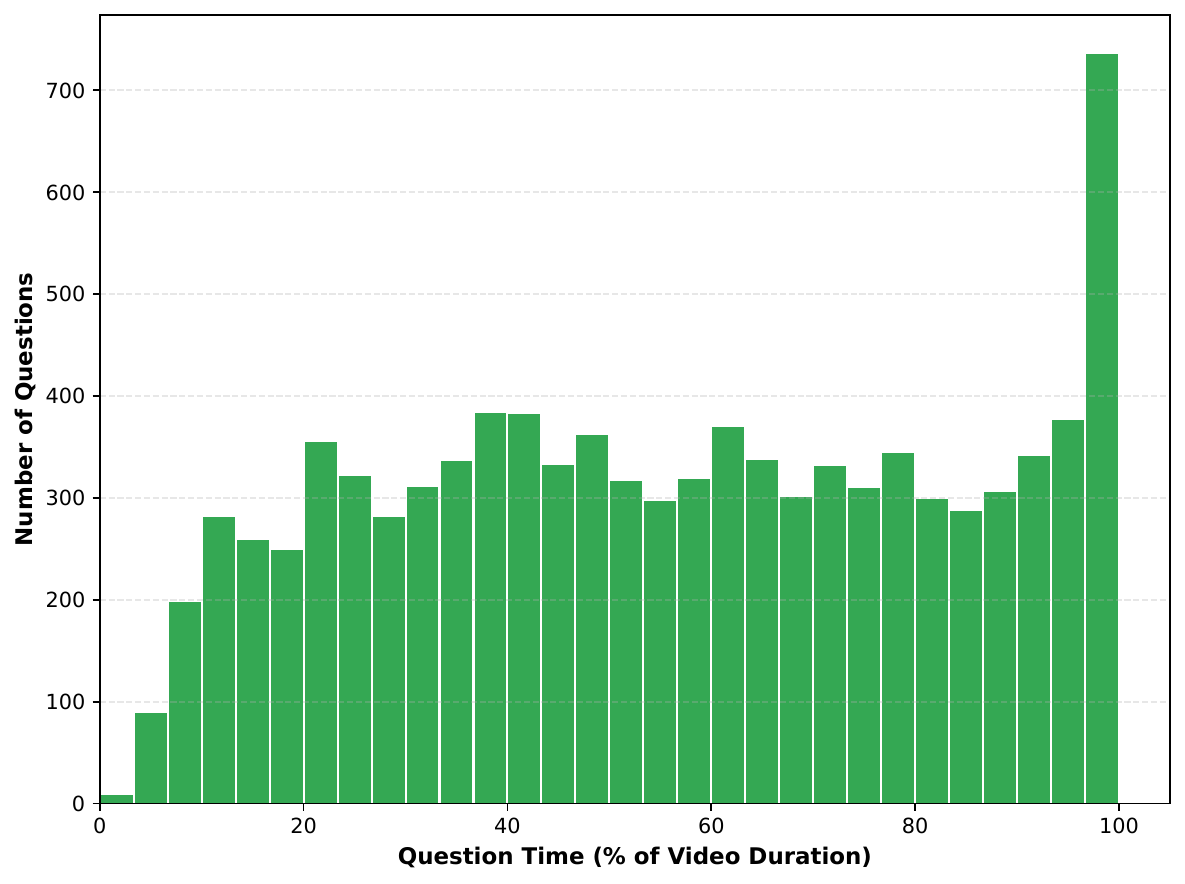}
      \caption{Question asking time is well distributed, requiring models to recall information whether recent or distant.}
      \label{fig:stats_time}
  \end{subfigure}
  \hfill
  \begin{subfigure}[b]{0.49\textwidth}
      \centering
      \includegraphics[width=0.9\textwidth]{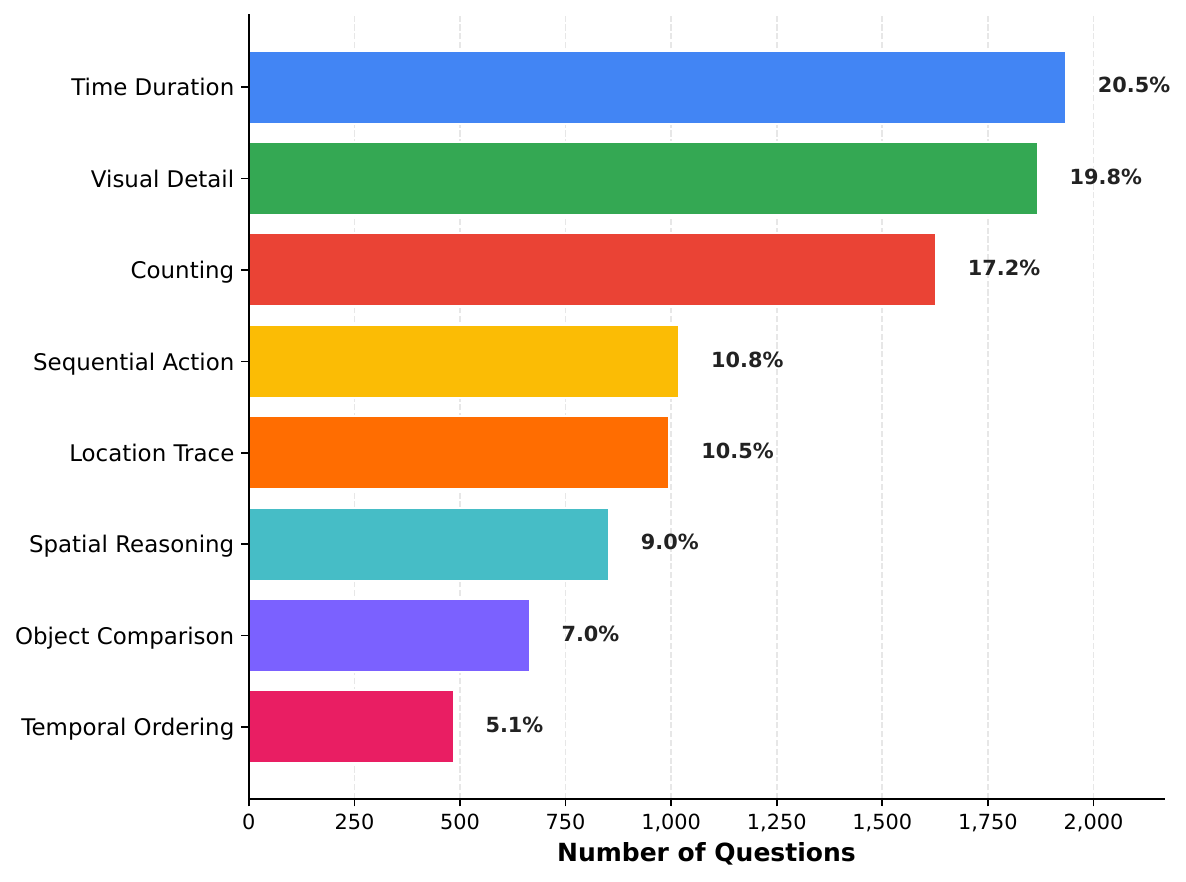}
      \caption{Questions span eight types, some requiring processing long video segments for counting and duration tasks.}
      \label{fig:stats_cat}
  \end{subfigure}
  \caption{\textbf{Data Statistics.}}
  \label{fig:stats}
\end{figure}

Questions are distributed approximately uniformly across the video timeline (Figure~\ref{fig:stats_time}), with a moderate skew toward the latter portion. This slight end-heaviness reflects the inclusion of summary-style questions that naturally require viewing the full video, while earlier questions tend to target specific moments or objects.

\section{Two Evaluation Tasks and Metrics}
\label{tasks}

\subsection{Streaming Answer Generation}
In this task, a model is provided with a video stream $V$ truncated at the ``natural trigger'' timestamp $T$. The model needs to generate a natural language response based solely on the historical context $V_{[0, T]}$. 

\textbf{Free-form Semantic Evaluation}: 
We utilize an LLM-as-a-Judge framework to compare the model’s prediction against our tri-tiered ground-truth answers (Surgeon, Chef, and Architect). A prediction is marked correct if it maintains semantic alignment with any of the three reference tiers. To provide a nuanced understanding of model reliability, we report two metrics: overall accuracy ($Acc_{ov}$), which penalizes only explicit counterfactuals to reward high-fidelity recall, and clean accuracy ($Acc_{cl}$), a strict metric that penalizes any hallucinated information regardless of its factual validity. We posit that the ``true'' accuracy of a model resides within the interval defined by these two bounds. Additional details on the LLM judge validation and the hallucination computation logic are provided in Appendix \ref{App:judge} and \ref{App:hall}.

\textbf{5-way Multiple Choice (MCQ)}: To provide a standardized, reproducible baseline that eliminates potential judge bias, we introduce a 5-way MCQ variant. The distractors are synthetically generated using an LLM, conditioned on both the ground-truth answers and a collection of hard negatives---incorrect predictions harvested from our baseline generative models. This ensures that the MCQ variant remains challenging and resistant to simple linguistic elimination.

\subsection{Temporal Grounding Retrieval}

Beyond semantic accuracy, a proactive assistant should justify its responses with visual evidence. In this task, the model predicts the evidential time interval $[t_{start}, t_{end}]$ supporting its answer. We evaluate the predicted interval against manually annotated ground-truth segments using three standard metrics:

\begin{itemize}
    \item Mean Intersection over Union (mIoU): We measure the average temporal overlap between predicted and ground-truth intervals across all queries. This serves as the primary metric for localization precision, following standard practices in temporal action localization~\citep{shou2016cdc}.
    \item Recall@1 ($IoU \ge \tau$): We report the percentage of predicted intervals that achieve a temporal $IoU$ of at least $\tau$ with the ground truth, regardless of the correctness of the generated text. This follows standard protocols in natural language video localization~\citep{gao2017tall, grauman2022ego4d}.
    \item Grounded QnA (GQ@$\tau$): To strictly penalize temporal hallucinations, we adopt the GQ@$\tau$ metric introduced by NExT-GQA~\citep{xiao2024nextgqa}. This joint metric requires a model to provide both a correct semantic answer and a precise temporal interval. 
    Unlike $R@1$, which measures grounding in isolation, GQ@$\tau$ ensures that a model is only credited when it reaches the correct conclusion based on the correct temporal evidence. 
\end{itemize}

\begin{table}[t]
\centering
\caption{Model performance comparison. Open-source models are evaluated at a matched budget of 128 uniformly sampled frames; proprietary models are subject to their platforms' own limits (Appendix~\ref{App:sampling}).}
\label{tab:model_comparison}
\small
\begin{tabularx}{\textwidth}{@{} l *{7}{Y} @{}}
\toprule
& \multicolumn{3}{c}{\textbf{Free form questions}} & \textbf{MCQ} & \multicolumn{3}{c}{\textbf{Interval Prediction}} \\
\cmidrule(r){2-4} \cmidrule(lr){5-5} \cmidrule(l){6-8}
\textbf{Model} & \textbf{$Acc_{cl}$} & \textbf{$Acc_{ov}$} & Halluc. & \textbf{$Acc$} & \textbf{$mIoU$} & R@1$\ge$.5 & GQ@.5 \\
\midrule
Human Ceiling & 91.43\% & 91.43\% & 0.00\% & -- & 0.777 & 0.858 & 0.698 \\
Gemini 3.1 Pro & 42.88\% & 47.83\% & 40.17\% & 54.41\% & 0.479 & 0.503 & 0.297 \\
InternVL3.5-38B & 25.52\% & 25.68\% & 25.22\% & 36.36\% & 0.244 & 0.220 & 0.082 \\
Qwen3VL-32B & 24.29\% & 30.32\% & 46.12\% & 42.40\% & 0.412 & 0.427 & 0.163 \\
GPT-5.4 & 23.77\% & 37.92\% & 58.90\% & 41.82\% & 0.333 & 0.313 & 0.117 \\
Socratic Baseline & 23.04\% & 32.25\% & 54.36\% & 31.21\% & 0.393 & 0.406 & 0.129 \\
Llava-OneVision-7B & 16.60\% & 18.38\% & 46.62\% & 30.53\% & 0.105 & 0.078 & 0.006 \\
Blind LLM & 2.68\% & 5.18\% & 82.47\% & 20.17\% & -- & -- & -- \\
\bottomrule
\end{tabularx}
\end{table}

\section{Evaluation of SOTA Solutions}
\label{experiments}

\subsection{Baselines}
\paragraph{Blind LLM.}
To quantify the ``vision-tax'' of the benchmark and establish a performance floor, we evaluate a text-only LLM (GPT-4o~\citep{openai2024gpt4o}) provided with only the question and the temporal anchor metadata, withholding all visual input. This baseline serves as a diagnostic tool to measure the extent to which linguistic statistical priors or common-sense reasoning can predict the correct answer. A low baseline performance serves as a sanity check, indicating that the dataset requires genuine multimodal reasoning to achieve high accuracy.

\paragraph{Socratic Video Reasoning.}
We employ a Socratic reasoning pipeline that discretizes the video into a structured textual history. This modular approach involves sampling frames at a rate of $0.2$ fps from the start of the video until the trigger timestamp and generating dense per-frame descriptions using a high-fidelity image captioner (GPT-4o). These sequential captions are then concatenated into a single prompt for a long-context LLM to answer the episodic question, testing whether textual synthesis can substitute for native multimodal integration.

\paragraph{Native Multimodal LLMs (MLLMs).}
We evaluate both closed-source and open-source multimodal models that process video natively. Our closed-source evaluation includes frontier models such as Gemini 3.1 Pro~\citep{geminiteam2025gemini3}, which supports extensive video context, alongside GPT-5.4~\citep{singh2025gpt5card}, which utilize high-resolution image-sequence reasoning. For the open-source community, we include representative models like InternVL3.5~\citep{wang2025internvl35}, Qwen3VL~\citep{bai2025qwen3vl}, and LLaVA-OneVision~\citep{li2024llavaonevision} to benchmark current capacity for long-form egocentric reasoning.

\paragraph{Human Performance.}
We establish a human reference directly from the annotations: semantic accuracy is the
proportion of independently written answers verified as correct, and the grounding ceiling is the agreement between annotators' intervals for the same question. Because boundary judgments vary between annotators, this ceiling is $0.777$ mIoU rather than $1.0$.

\begin{figure}[tbp]
     \centering
     \scalebox{0.9}{
     \begin{subfigure}[b]{0.48\textwidth}
         \centering
         \includegraphics[width=\textwidth]{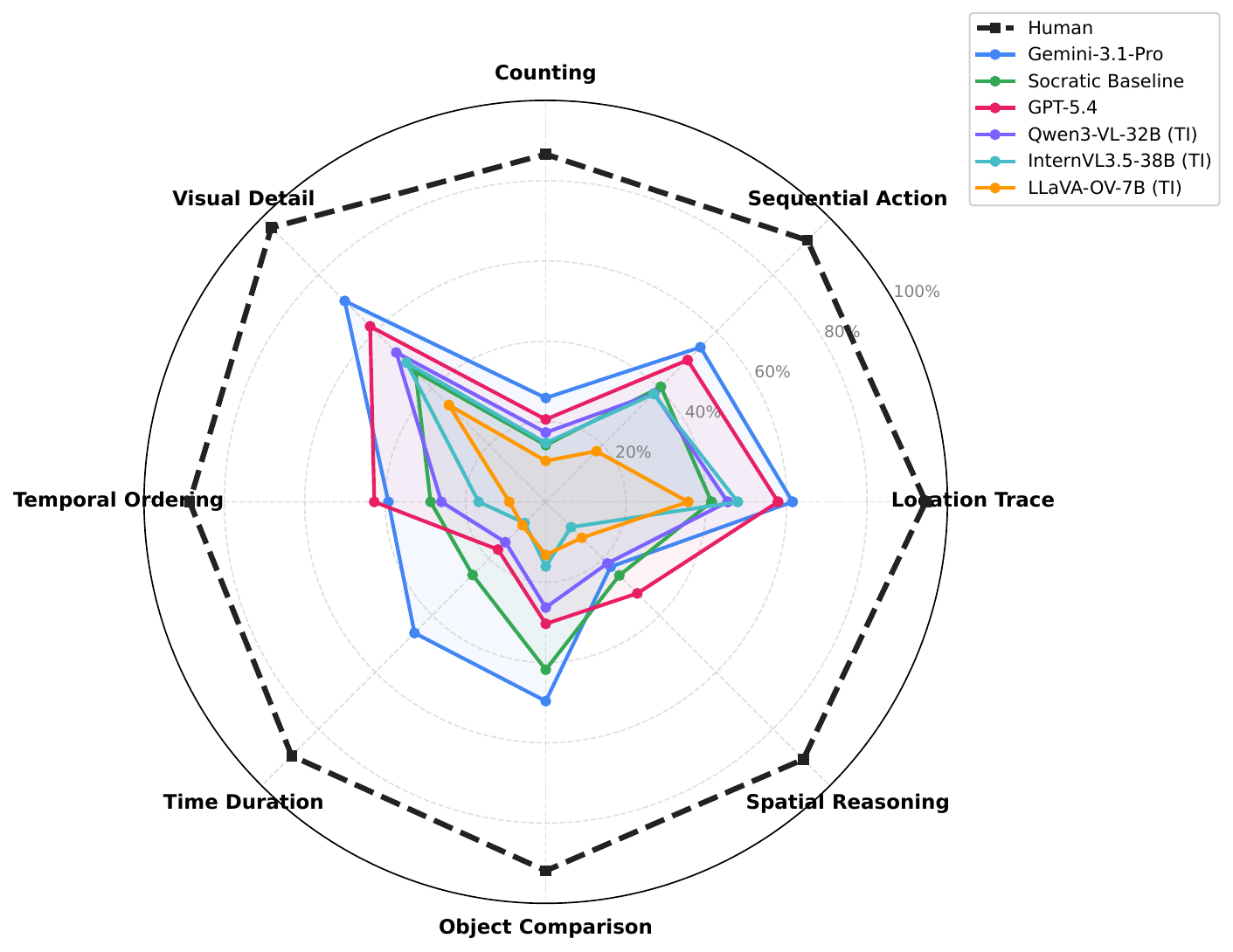}
         \caption{Accuracy by Category}
         \label{fig:radar}
     \end{subfigure}
     \hfill 
    \begin{subfigure}[b]{0.45\textwidth}
         \centering
         \includegraphics[width=\textwidth]{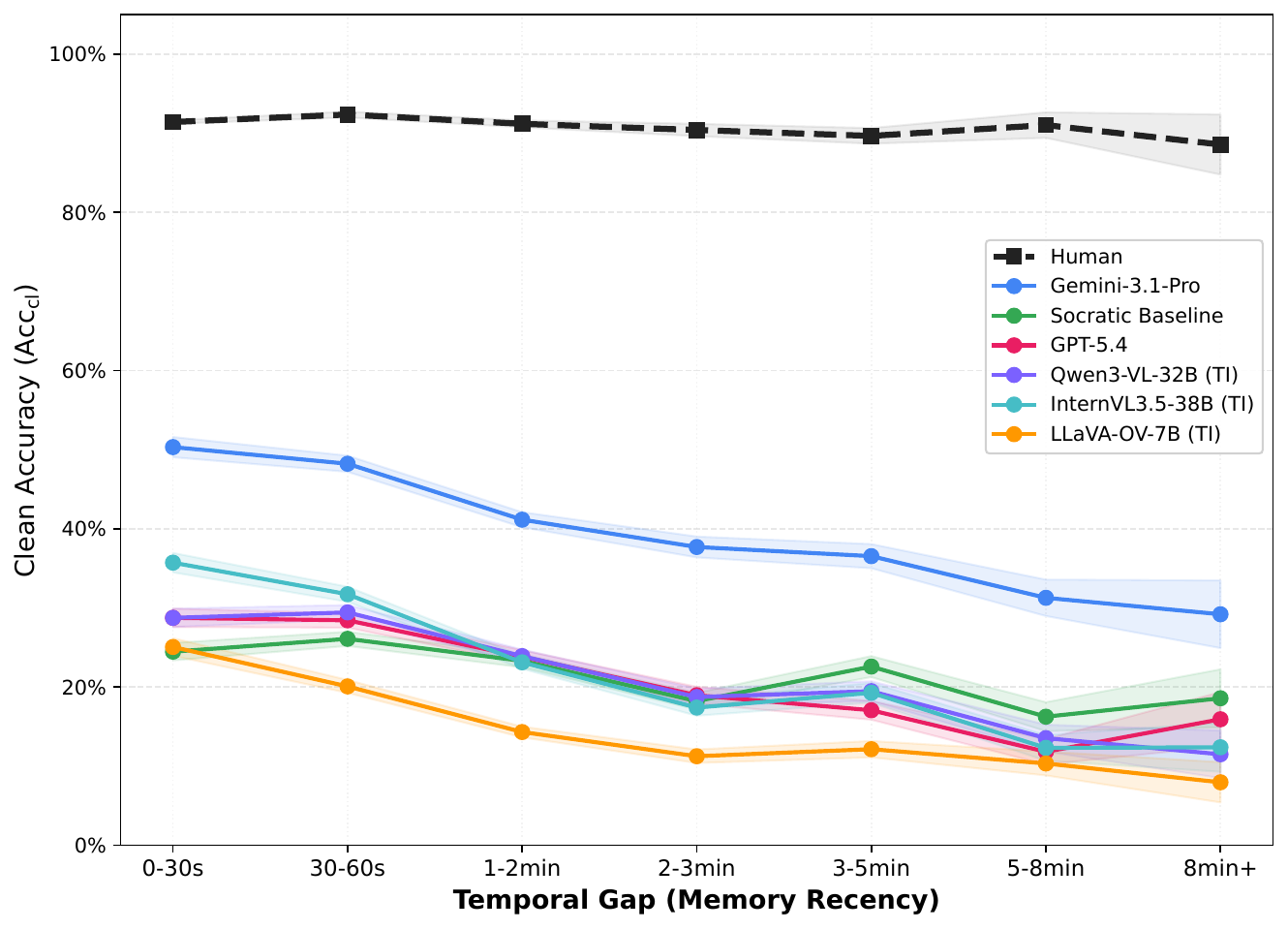}
         \caption{Accuracy vs. Memory Recency \\
         Shaded bands denote $\pm 1$ standard error.}
         \label{fig:memory_decay}
     \end{subfigure}
     }
     \caption{\textbf{Comparative Analysis of Model Performance.}}
     \label{fig:combined-analysis}
\end{figure}

\subsection{Results}
\label{sec:results}

\paragraph{Semantic Reasoning.}
Table \ref{tab:model_comparison} summarizes the evaluation results on S-EMBER. Frontier models demonstrate clear leadership in reasoning-centric metrics, with Gemini 3.1 Pro achieving the highest overall performance in both Clean Accuracy (42.88\%) and 5-way MCQ Accuracy (54.41\%). Image-sequence models such as GPT-5.4 and open-source video models follow, clustering between 23\% and 26\% Clean Accuracy. The significant delta between Clean Accuracy (which penalizes any unverified info) and Overall Accuracy (which allows for orthogonal details) underscores the pervasive challenge of balancing high-fidelity recall with strict factual grounding.


\paragraph{Temporal Grounding Bottleneck.} Gemini 3.1 Pro leads on localization at $0.479$ mIoU. Among the open-source models, evaluated at an identical 128-frame budget, Qwen3VL-32B reaches $0.412$, ahead of GPT-5.4 ($0.333$) and InternVL3.5-38B ($0.244$). The Socratic pipeline remains competitive at $0.393$ despite performing no video modelling whatsoever, which suggests that a substantial part of what current MLLMs lack for this task is an explicit temporal index rather than visual capability. Grounded accuracy, a correct answer drawn from a correctly located interval, reaches only $0.297$ for the best frontier model and $0.163$ for the best open model, against a human ceiling of $0.698$. Measured as a fraction of human performance, the strongest model attains $62\%$ of the human mIoU and $52\%$ of human accuracy but only $43\%$ of the human joint rate, indicating that the two failure modes do not coincide: the models that locate the evidence are not consistently the ones that answer from it.

\paragraph{Reliability and Hallucination Analysis.} 
The Blind LLM baseline (GPT-4o), yielding 2.68\% Clean Accuracy and near-random MCQ Accuracy, confirms that S-EMBER successfully filters out linguistic shortcuts and common-sense priors. Among successful models, we observe a wide variance in reliability. InternVL3.5 demonstrates the lowest hallucination rate (25.22\%) among high-performing models, suggesting a more conservative and reliable retrieval strategy. In contrast, Qwen3VL and Gemini 3.1 Pro exhibit higher hallucination rates (46.12\% and 40.17\%, respectively), often generating plausible but unverified details. The high hallucination rates in the Socratic baselines (up to 54.36\%) highlight a ``synthesis failure'': while modular captioning excels at local grounding, the subsequent LLM stage frequently struggles to synthesize these textual descriptions without introducing linguistic noise.

\subsection{Diagnostic Analysis}
\label{sec:diagnostics}

\paragraph{Performance by Category.}
The radar chart in Figure \ref{fig:radar} illustrates the breakdown of semantic accuracy across the eight question categories. While frontier models like Gemini 3.1 Pro and GPT-5.4 exhibit relatively balanced profiles, they still fall significantly short of the human ceiling, particularly in \textbf{Counting and Spatial Reasoning}, which require persistent state tracking over long horizons. Notably, Gemini 3.1 Pro demonstrates a substantial lead in Time Duration over other native models, showcasing a superior capacity for reasoning across long-form video signals.


\paragraph{Memory Recency.} We observe a \textbf{universal recall decay across all models} as the temporal gap between the evidence and the query increases (Figure \ref{fig:memory_decay}). Human performance is essentially flat, between 88\% and 93\% at every gap, so the decline is a property of the models rather than of the questions. Gemini 3.1 Pro falls steadily from 50.3\% for immediate recall (0--30s) to 29.2\% beyond eight minutes; the open-source models decline faster from a lower starting point, Qwen3-VL-32B from 28.7\% to 11.5\% and InternVL3.5-38B from 35.7\% to 12.4\% over the same range. The Socratic baseline decays most gracefully in relative terms, retaining 76\% of its initial accuracy at the longest gaps against Qwen3-VL's 40\%, which suggests that explicit textual logs are a less volatile store than the internal temporal buffers of end-to-end transformers over long horizons.

\section{Ablation Study on Model Scale, Temporal Density, and Visual Fidelity}
\label{sec:ablation}

\begin{figure}[tbp]
  \centering
  \begin{subfigure}[b]{0.31\textwidth}
      \centering
      \includegraphics[width=\textwidth]{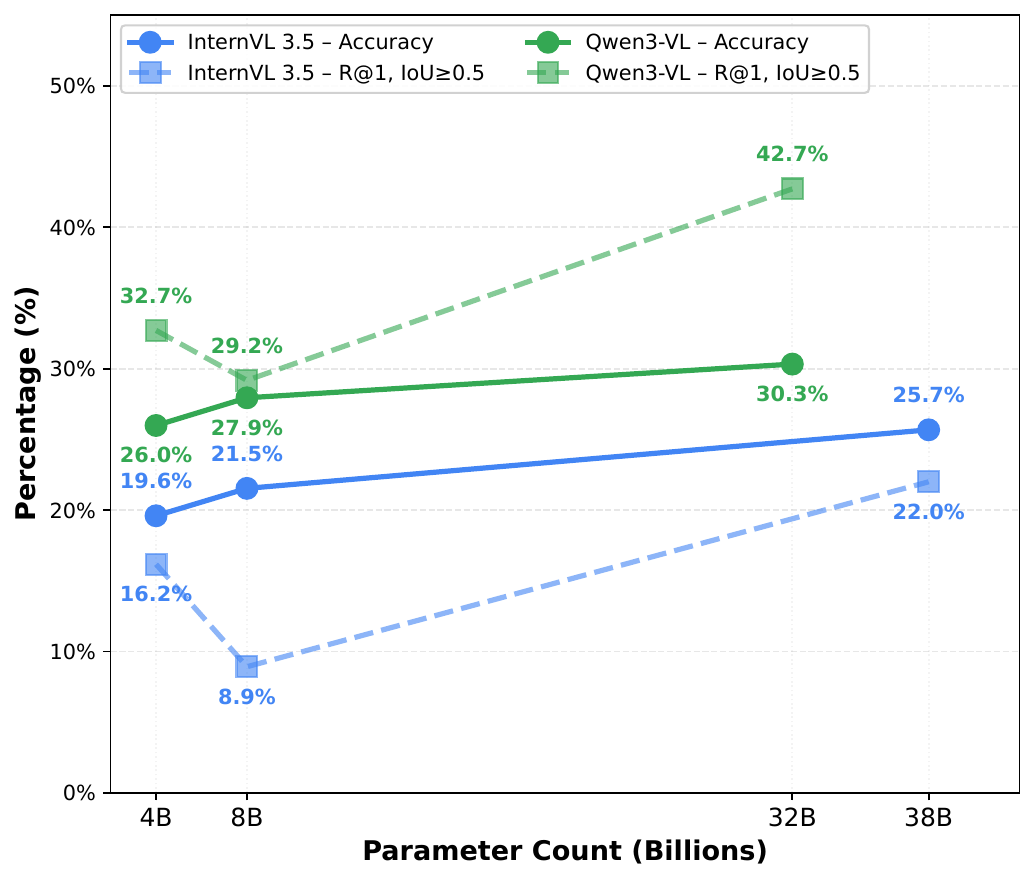}
      \caption{Model Scaling}
      \label{fig:scale}
  \end{subfigure}
  \hfill
  \begin{subfigure}[b]{0.31\textwidth}
      \centering
      \includegraphics[width=\textwidth]{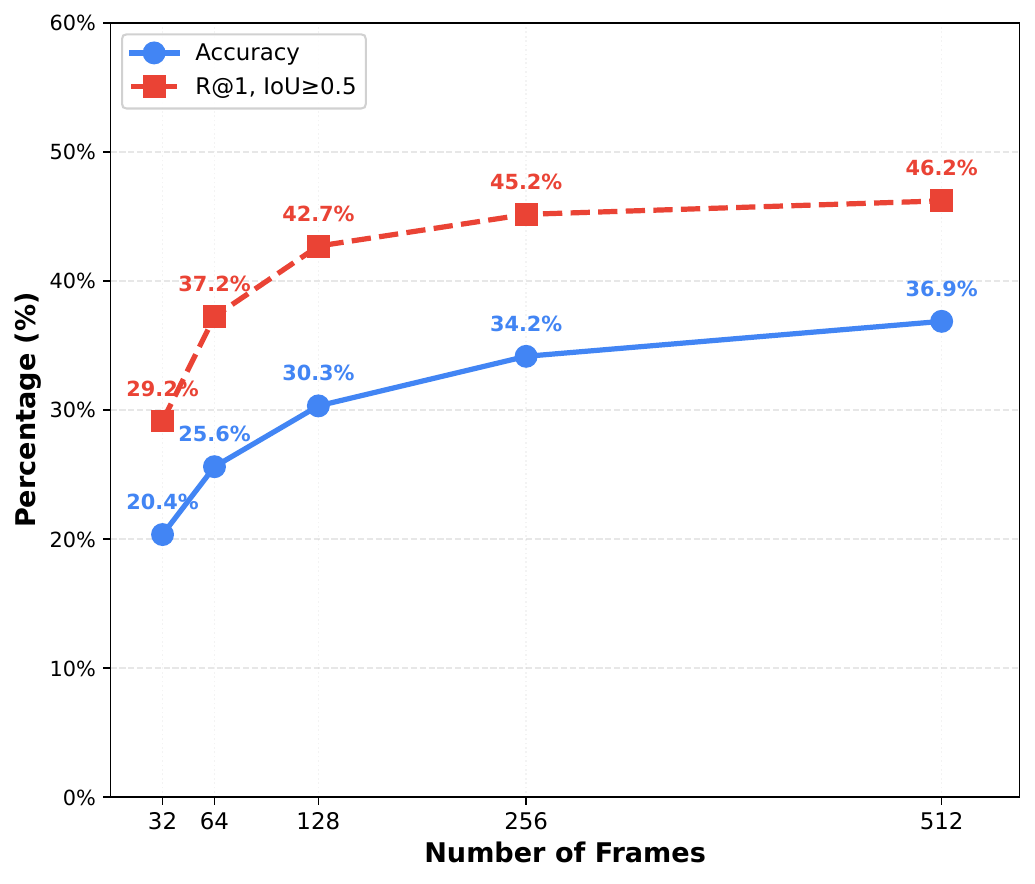}
      \caption{Frame Count Ablation}
      \label{fig:frame}
  \end{subfigure}
  \hfill
  \begin{subfigure}[b]{0.31\textwidth}
      \centering
      \includegraphics[width=\textwidth]{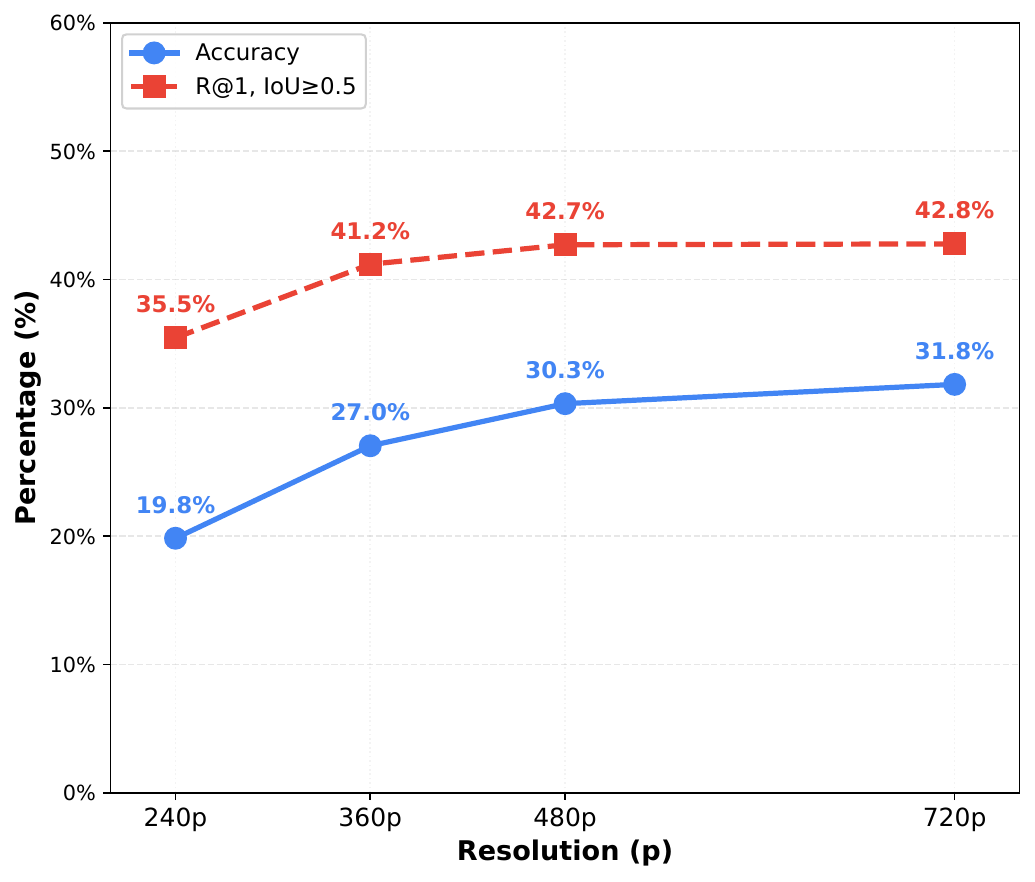}
      \caption{Resolution Ablation}
      \label{fig:res}
  \end{subfigure}
  \caption{\textbf{Comprehensive Model Evaluation.}}
  \label{fig:triple-comparison}
\end{figure}

\paragraph{Impact of Model Scale.} We evaluate InternVL3.5 at 4B/8B/38B and Qwen3VL at 4B/8B/32B at a matched budget of 128 frames (Figure~\ref{fig:triple-comparison}a). Semantic accuracy scales monotonically in both families, from 19.6\% to 25.7\% and from 26.0\% to 30.3\% respectively. Grounding does not: in both families the 8B model localizes \emph{worse} than its 4B counterpart, 29.7 against 32.8 mIoU for Qwen3VL and 13.2 against 20.4 for InternVL3.5, even as its accuracy and multiple-choice scores improve. Parameter count therefore purchases a better answer without reliably purchasing a better location for it, which is difficult to reconcile with the view that localization is simply a more demanding function of the same underlying perception.


\paragraph{Temporal Density vs. Reasoning Accuracy.} We investigate the effect of frame sampling rate using Qwen3VL-32B by varying the frame count from 32 to 512 at fixed resolution (Figure~\ref{fig:triple-comparison}b). Accuracy grows logarithmically, from 20.4\% to 36.9\%, underscoring the ``needle-in-a-haystack'' nature of S-EMBER: dense sampling is required simply to capture the brief visual anchors on which recall depends. Grounding improves alongside it, from 30.7 to 44.4 mIoU, but with returns that flatten considerably sooner---increasing the budget from 32 to 128 frames contributes $+10.5$ mIoU, whereas quadrupling it again to 512 adds only $+3.2$. Extrapolating this trend, reaching the human ceiling of 77.7 by density alone would demand a frame budget far beyond what current architectures can attend over. Denser sampling is necessary for temporal grounding, but it is not sufficient.


\paragraph{Resolution and the Reasoning Ceiling.} Finally, we ablate visual resolution by fixing the frame count at 128 and scaling the resolution from 240p to 720p (Figure~\ref{fig:triple-comparison}c). Higher fidelity significantly boosts accuracy, from 19.8\% to 31.8\%, likely due to improved recognition of fine-grained details and OCR-heavy evidence. Grounding behaves differently: it rises from 35.4 to 40.1 mIoU between 240p and 360p and then saturates, gaining a further $1.3$ mIoU across the subsequent fourfold increase in pixel count. The resolution required to localize an event is thus considerably lower than the resolution required to interpret it. Once a frame is legible enough to be placed on the timeline, additional visual clarity helps the model understand the evidence without helping it determine when that evidence occurred.

\section{Discussion: Pathways to Grounded Episodic Intelligence}
\label{Discussion}
 
Based on our diagnostic and ablation studies, we identify three critical areas for future development.

\textbf{Bridging the Localization-Reasoning Disconnect.}
Our results consistently show that models like GPT-5.4 and Gemini 3.1 Pro can often answer correctly while failing to locate the supporting evidence (\S\ref{sec:results}). This ``correct for the wrong reasons'' behavior indicates that current multimodal architectures rely too heavily on global scene context and linguistic priors. Future research should prioritize architectures that explicitly link generated tokens to temporal indices, perhaps through cross-modal attention mechanisms that enforce stricter alignment between the textual response and the visual proof interval.

\textbf{Explicit Temporal Metadata.} A model can report a timestamp only if its input establishes what the timestamps are, and video interfaces differ in whether they provide this. The effect is large in our experiments: supplying per-frame offsets to Qwen3VL-32B at a matched 128-frame budget raises grounding from $0.018$ to $0.412$ mIoU---a greater effect than scaling the model eightfold or increasing the frame count sixteenfold. Architectures differ in how far they exploit it, and LLaVA-OneVision, which provides no mechanism for binding a textual offset to a visual token, is unaffected ($0.104$ to $0.105$). Temporal grounding scores are therefore a property of the model and its input interface jointly, and we evaluate all open-source models under a single convention (Appendix~\ref{App:sampling}).


\textbf{Adaptive Temporal Indexing.} Grounding returns diminish well before the human ceiling (\S\ref{sec:ablation}), raising the question of whether the gap concerns \emph{how many} frames a model sees or \emph{which} ones. We evaluate a two-pass baseline that concentrates half of a fixed 128-frame budget around an interval predicted by a first pass, alongside an oracle variant given the ground-truth interval (Appendix~\ref{App:adaptive}). Adaptive indexing improves every metric for both model families and both task formats (Table~\ref{tab:adaptive}), including multiple choice, which produces no temporal output, so the gain is in answering rather than interval formatting. Yet the oracle, an upper bound on any policy for deciding where to look, reaches only $22.8$ GQ@0.5 against a human $69.8$. Frame selection is therefore not the binding constraint: under the oracle the residual shortfall lies mainly in answering rather than in localization.

\begin{table}[t]
\centering
\caption{Adaptive temporal indexing at a matched 128-frame budget.}
\label{tab:adaptive}
\small
\begin{tabularx}{\textwidth}{@{} l l *{7}{Y} @{}}
\toprule
& & \multicolumn{3}{c}{\textbf{Free form questions}} & \textbf{MCQ} & \multicolumn{3}{c}{\textbf{Interval Prediction}} \\
\cmidrule(lr){3-5} \cmidrule(lr){6-6} \cmidrule(l){7-9}
\textbf{Model} & \textbf{Frames} & \textbf{$Acc_{cl}$} & \textbf{$Acc_{ov}$} & Halluc. & \textbf{$Acc$} & \textbf{$mIoU$} & R@1$\ge$.5 & GQ@.5 \\
\midrule
Qwen3VL-32B     & Uniform  & 24.3 & 30.3 & 46.1 & 42.4 & 41.2 & 42.7 & 16.3 \\
                & Adaptive & 25.3 & 31.6 & 46.3 & 44.2 & 42.2 & 44.2 & 18.1 \\
                & Oracle   & 30.3 & 38.4 & 43.1 & 50.0 & 49.0 & 51.6 & 22.8 \\
\midrule
InternVL3.5-38B & Uniform  & 25.5 & 25.7 & 25.2 & 36.4 & 24.4 & 22.0 &  8.2 \\
                & Adaptive & 26.4 & 26.6 & 24.7 & 38.1 & 25.7 & 23.4 &  9.4 \\
                & Oracle   & 30.6 & 30.8 & 20.8 & 44.3 & 31.4 & 30.0 & 12.9 \\
\bottomrule
\end{tabularx}
\end{table}


\section{Conclusion and Limitations}
\label{limitation}
\label{conclusion}
In this work, we introduced S-EMBER, a large-scale egocentric benchmark designed to evaluate episodic memory within continuous, unscripted video streams. Using data captured via modern wearable smart glasses, S-EMBER provides an ecologically valid diagnostic for the next generation of always-on AI assistants. 
Through our episodic memory taxonomy and tri-tiered response system, we exposed a critical grounded recall gap: current MLLMs answer and localize with moderate competence in isolation but rarely achieve both on the same query, a deficit that scaling does not close. These findings establish a new frontier for the development of grounded streaming architectures, ensuring that future AI assistants can not only reason about our past but also precisely anchor that reasoning in the visual reality of our daily lives.

\textbf{Limitations:} A primary constraint of S-EMBER is its single-modality nature. By permanently redacting audio streams for modality isolation, we prevent models from utilizing acoustic cues that are often vital for real-world AI assistants. Consequently, S-EMBER does not evaluate the multi-modal integration for wearable intelligence. Additionally, although our LLM-as-a-Judge framework demonstrates high correlation with human experts in Appendix \ref{App:judge}, it remains an approximation that may exhibit verbosity bias or miss subtle nuances identified by human raters. Although our multi-tiered response system and manual audits mitigate these effects, the potential for minor evaluation artifacts persists. Finally, S-EMBER's coverage reflects organic, consent-constrained capture rather than a designed distribution: everyday routines such as shopping and food preparation dominate, and while questions are spread across the eight cognitive categories, the residual imbalance (largest $\approx 20\%$, smallest $\approx 5\%$) means per-category results rest on unequal sample sizes.

\section*{Acknowledgments}
We thank Akinniyi Akinyemi, Jiemin Zhang, and Somya Jain;
Christine Chen and Cynthia Gao; Sagar Miglani; and
Justine Kao, Maddie Mintz, and Vanessa Stark
for their contributions and support throughout this project.

\medskip
\bibliographystyle{plainnat}
\bibliography{paper}

\clearpage
\newpage
\beginappendix

\section{Additional Experimental Study and Figures}

\begin{figure}[htbp]
     \centering
     \begin{subfigure}[b]{0.48\textwidth}
         \centering
         \includegraphics[width=\textwidth]{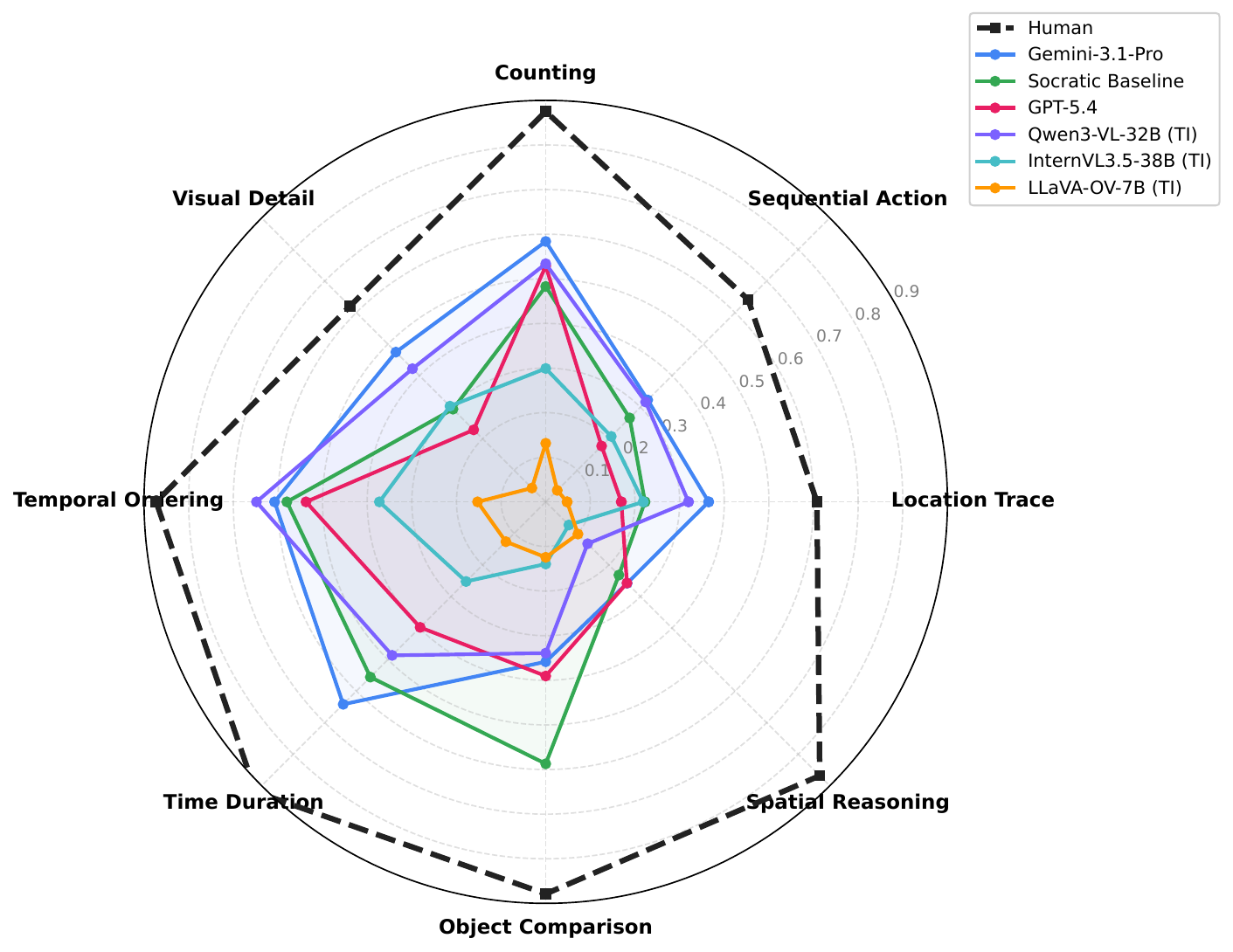}
         \caption{Temporal Grounding (mIoU) by Category}
         \label{fig:mIoU-bar}
     \end{subfigure}
     \hfill 
     \begin{subfigure}[b]{0.45\textwidth}
         \centering
         \includegraphics[width=\textwidth]{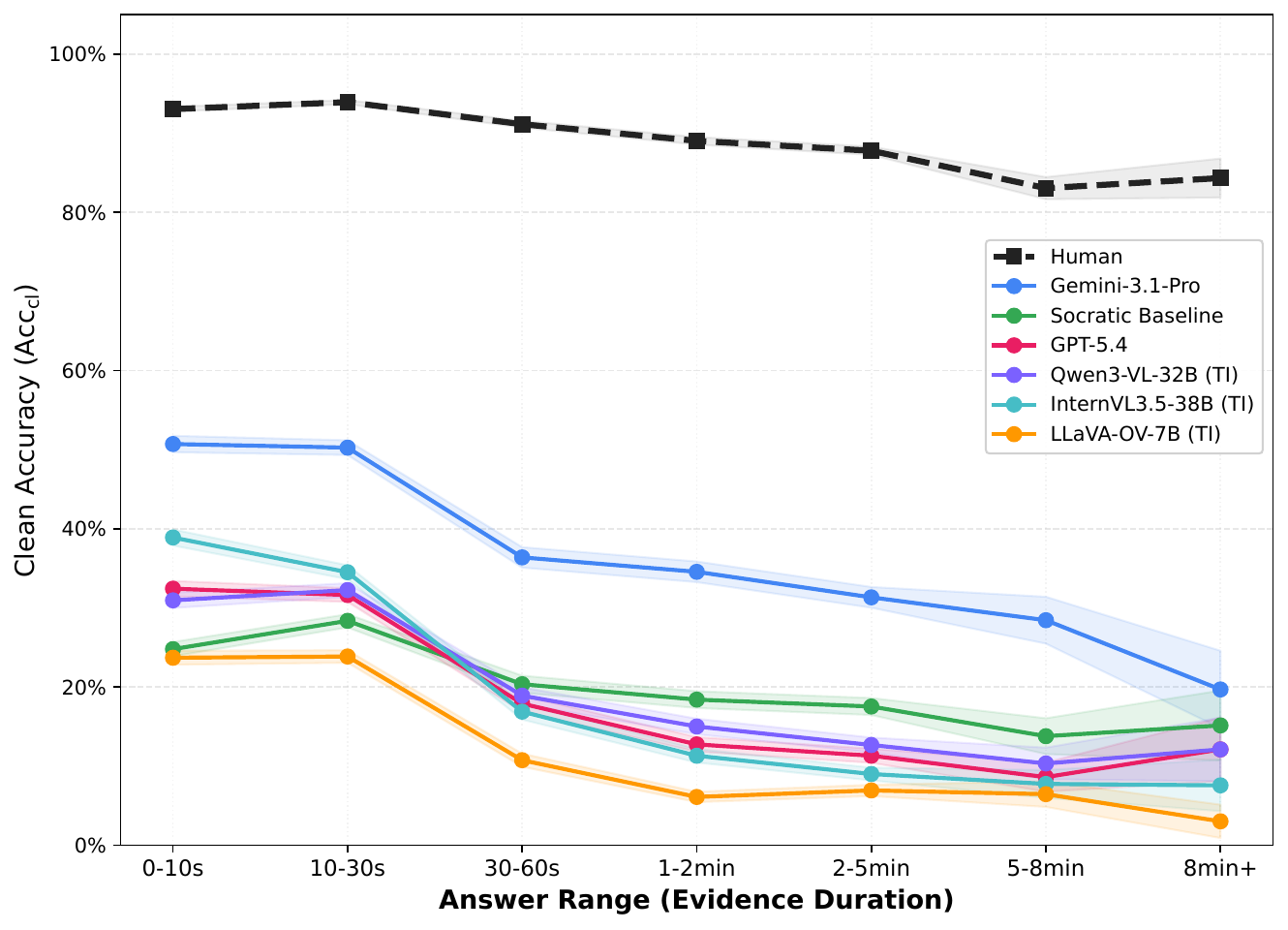}
         \caption{Accuracy vs. Answer Range}
         \label{fig:answer_range}
     \end{subfigure}
     \caption{\textbf{Accuracy Analysis of Model Performance.}}
\end{figure}

\textbf{Grounding by Category.}
Figure~\ref{fig:mIoU-bar} breaks temporal grounding down across the eight categories. Every model performs best where the timeline is explicit and linear (Temporal Ordering, Counting and Time Duration), with Qwen3VL-32B reaching $0.648$ mIoU and overtaking Gemini 3.1 Pro ($0.608$). Performance falls sharply on the spatially driven categories: Spatial Reasoning is the weakest for every model, at $0.133$ for Qwen3VL against a human ceiling of $0.868$, making the anchoring of spatial transitions to a temporal index the hardest demand S-EMBER poses. The Socratic baseline is not uniformly behind, its dense textual index yielding the highest Object Comparison grounding we measure ($0.587$). One model does collapse outright: LLaVA-OneVision-7B remains at or below $0.15$ mIoU in all eight categories.

\textbf{Evidence Duration.} 
The semantic accuracy scales inversely with the duration of the visual evidence (Figure \ref{fig:answer_range}). Models are most effective when the evidence is brief, within the first half-minute, likely because short intervals act as distinct visual ``needles.'' However, as evidence becomes more continuous---requiring the synthesis of multiple minutes of video---accuracy drops sharply for all native multimodal models. This highlights a persistent long-form synthesis bottleneck: while models can identify isolated snapshots, they \textbf{struggle to aggregate information across extended temporal ranges} to form a single coherent answer.

\begin{figure}[ht]
     \centering
     \includegraphics[width=0.90\textwidth]{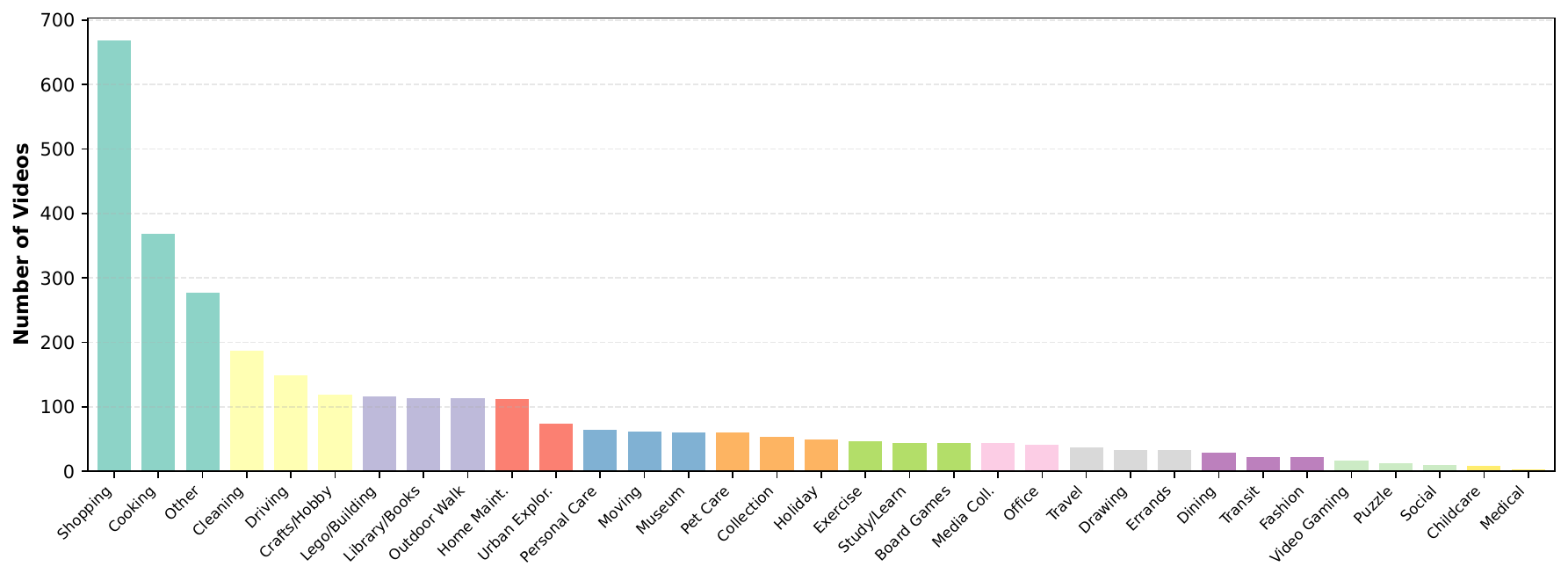}
     \caption{Video Scenario Distribution.}
     \label{fig:scenario}
\end{figure}

\section{LLM Judge Validation}
\label{App:judge}

To validate the reliability of our LLM judge (Gemini 3.1 Flash), we constructed a controlled meta-evaluation set from our human-verified annotations. The goal was to test whether the judge could accurately distinguish between correct and incorrect model outputs when compared against a "gold standard" reference.

\textbf{Data Construction:} We selected a subset of questions that contained two distinct correct human-provided answers ($C_1, C_2$) and one incorrect answer ($W$). From these, we built 600 evaluation pairs (300 positive and 300 negative) using a "leave-one-out" strategy.
\begin{itemize}
    \item Positive Samples (Correctness Check): We present the judge with one correct answer ($C_1$) as the "model prediction" and use the other correct answer ($C_2$) as the ground-truth reference. A reliable judge should mark these as a match.
    \item Negative Samples (Discrimination Check): We present the judge with a verified incorrect answer ($W$) as the "model prediction" against the correct reference ($C_2$). A reliable judge should reject these.
\end{itemize}

\textbf{Validation Results:} The validation results demonstrate that the judge is both highly accurate and appropriately skeptical. It achieved an overall accuracy of 96.2\%, with 97.2\% precision and 95.7\% recall ($F_1 = 0.964$). Critically for benchmark integrity, the False Positive Rate (FPR) was only 1.5\%. This indicates that the judge is extremely conservative, very rarely awarding credit to incorrect answers. These metrics confirm that our LLM-based evaluation pipeline serves as a rigorous and reliable proxy for human judgment in open-ended VideoQA scoring.

\textbf{Per-category reliability.} We also report reliability by cognitive category
(Table~\ref{tab:judge_percat}). Accuracy exceeds $85\%$ and precision exceeds $92\%$ in
every category, including the open-ended spatial and trace categories in which model
responses are longest. Precision exceeds accuracy throughout, so the conservatism noted
above holds category by category rather than only in aggregate.

\begin{table}[h]
\centering
\caption{LLM judge reliability by cognitive category, on 600 human-verified pairs.}
\label{tab:judge_percat}
\small
\begin{tabularx}{0.75\textwidth}{@{} l *{3}{Y} @{}}
\toprule
\textbf{Cognitive category} & \textbf{Pairs ($n$)} & \textbf{Accuracy} & \textbf{Precision} \\
\midrule
Counting objects/events & 158 & 100.0\% & 100.0\% \\
Time / duration         & 90  & 97.8\%  & 100.0\% \\
Visual detail recall    & 80  & 97.5\%  & 95.3\% \\
Sequential action       & 60  & 96.7\%  & 97.1\% \\
Object comparison       & 48  & 95.8\%  & 96.9\% \\
Temporal ordering       & 44  & 95.5\%  & 96.4\% \\
Spatial-aware reasoning & 72  & 91.7\%  & 93.9\% \\
Location trace          & 48  & 85.4\%  & 92.3\% \\
\midrule
\textbf{Overall}        & \textbf{600} & \textbf{96.2\%} & \textbf{97.2\%} \\
\bottomrule
\end{tabularx}
\end{table}

\section{Hallucination Logic and Metric Computation}
\label{App:hall}

To distinguish reliable recall from fabrication, the LLM-as-a-Judge evaluates the model's prediction for hallucinated information by checking the response against the union of all three ground-truth tiers (Surgeon, Chef, and Architect). A prediction is flagged as hallucinating if it introduces any extrinsic details, entities, or assertions that are not explicitly supported by this reference set. This "union-check" ensures that a model is never penalized for providing a detail that was captured in our most comprehensive ground-truth tier (Architect) but omitted from the most concise one (Surgeon). Based on this hallucination audit, we compute our two primary semantic metrics as follows:
\begin{itemize}
    \item Overall Accuracy ($Acc_{ov}$): This metric measures high-fidelity recall. A prediction is marked correct if it aligns with the core factual requirements of the ground truth and contains no explicit counterfactuals (direct lies). Under this metric, a model is not penalized for being "talkative" or adding unverified details, provided those details do not contradict the known facts.
    \item Clean Accuracy ($Acc_{cl}$): This is a strict reliability metric. A prediction is only marked as "Clean" if it is semantically correct (as per $Acc_{ov}$) and contains zero hallucinated information. If a prediction introduces even a single extrinsic detail not found in the ground-truth union, it is disqualified from $Acc_{cl}$.
\end{itemize}

By reporting both metrics, we define a performance interval that exposes the gap between a model's ability to retrieve the correct answer and its tendency to embellish that answer with fabricated evidence.

\section{Video Processing and Frame Sampling Strategy}
\label{App:sampling}



For efficient downstream processing, all raw video footage is stored at a standardized
resolution of 720p at 12\,fps. Model inputs are sampled from this encoding; per-frame
resolutions quoted below and in \S\ref{sec:ablation} refer to the input presented to the
model rather than to the stored source.

To keep cross-model comparisons interpretable, every open-source model is evaluated under an identical input protocol: 128 uniformly sampled frames drawn from the streaming prefix $[0, t_{question}]$, each annotated with its true offset in the source video. Absolute time must be supplied for the interval-prediction task to be well posed, since without it a model can describe when an event occurred only relative to the frames it happens to have been shown. Two mechanisms deliver these offsets depending on the architecture: Qwen3VL accepts native \texttt{<t seconds>} timestamp tokens, while InternVL3.5 and LLaVA-OneVision receive a textual frame prefix of the form \texttt{Frame $i$ (at $t$s):}. Model weights and frame budgets are identical in either case. Per-frame resolution is likewise chosen for comparability rather than to maximise any one model: InternVL3.5 processes each frame as a single $448\times448$ tile, LLaVA-OneVision resizes frames with its own image processor, and Qwen3VL, which takes an explicit pixel budget, is set to 480p so that its per-frame budget falls in the same range. The frame-count and resolution ablations of \S\ref{sec:ablation} vary one axis at a time from this operating point.

Proprietary models cannot be brought onto the same pipeline. GPT-5.4 is capped at 50 images per request by the Azure platform, and Gemini 3.1 Pro accepts raw video files through a native API that performs its own sampling at 1\,fps and carries temporal positions internally. We therefore treat the proprietary results as reference points rather than controlled comparisons, and draw every diagnostic conclusion in \S\ref{sec:ablation} from within-model ablations, which are free of cross-model interface confounding.

\section{Experiments Compute}
\label{App:compute}
All open-source baselines were evaluated on a single node equipped with $8\times$ NVIDIA H200 GPUs. End-to-end inference times range from approximately $5.5$\,h for LLaVA-OneVision-7B and $\sim 11.5$\,h for InternVL3.5-4B/8B, up to $\sim 19.5$\,h for InternVL3.5-38B. 
Qwen3VL-32B at a fixed per-frame resolution scales from $\sim 4$\,h at 32 input frames
to $\sim 32$\,h at 768 frames; the primary 128-frame configuration completes in $\sim 8$\,h. The two-pass adaptive baseline costs approximately twice the single-pass 128-frame runtime, as the video is read once per pass.

\section{Two-Pass Adaptive Temporal Indexing}
\label{App:adaptive}
The adaptive baseline of \S\ref{Discussion} operates in two passes over the streaming
prefix $[0, t_{question}]$ and requires no retraining.

\textbf{Pass 1 (indexing).} The model samples 64 frames uniformly over the prefix and
answers the grounding prompt. Its answer text is discarded and only the predicted interval is retained, as a temporal index. 

\textbf{Pass 2 (answering).} The video is re-read with 128 frames: the same 64 coarse frames, plus 64 sampled uniformly inside a window centred on the predicted interval and widened by a factor $k$, merged into a single non-uniformly spaced timeline carrying true absolute timestamps. Both passes remain inside $[0, t_{question}]$, preserving the streaming constraint, and the total budget matches the uniform baseline exactly; only the placement of the frames differs.

\textbf{Oracle window.} The oracle variant is identical except that the first pass's
estimate is replaced by the ground-truth interval, widened by one coarse sampling step on either side. It is not a deployable method, but it bounds the performance of any policy for deciding where to look under a fixed frame budget.

\textbf{Sensitivity to $k$.} Table~\ref{tab:adaptive} reports $k=8$. The grounding gains
hold across $k \in \{4, 8, 20, 30\}$, so the result does not depend on a tuned constant, and bootstrap 95\% confidence intervals on the grounding deltas against the uniform control exclude zero at every setting.

\section{Annotation Protocol}
\label{App:protocol}

This section details the three-stage protocol used to develop S-EMBER benchmark, designed to evaluate temporal and spatial reasoning in long-form (5--20 minute) egocentric video.

\subsection*{Stage 1: Question Generation}
The first stage focuses on eliciting natural, perspective-based questions that challenge long-term memory. To ensure the benchmark tests sustained comprehension rather than immediate recall, we enforce a one-minute initial buffer during which no questions may be authored. For the remaining duration, annotators must maintain comprehensive coverage by sourcing at least one question from each subsequent third of the video. Furthermore, questions must span eight distinct memory categories, with a diversity requirement of at least three unique types per video. Crucially, all questions are authored with a temporal timestamp that strictly follows the appearance of the relevant visual evidence.

\subsection*{Stage 2: Answer Writing and Grounding}
In the second stage, annotators generate grounded answers and identify supporting visual intervals. To account for varying linguistic preferences, answers are produced at three Levels of Detail (LoD). These are defined by specific personas: \textbf{Surgeons} provide concise, direct answers; \textbf{Chefs} offer balanced responses with clarifying details; and \textbf{Architects} deliver comprehensive descriptions including all structural context. Each answer is paired with a temporal grounding interval. These intervals are governed by a strict padding rule, where the added buffer time must not exceed 15\% of the evidence duration or a maximum 15-second cap.

\subsection*{Stage 3: Quality Verification} 

Verification is sequential rather than parallel: each triplet is adjudicated by a single
reviewer, and disagreement with Stage 2 results in rejection rather than arbitration. Agreement before adjudication is nonetheless high. Independent groundings of the same question have a mean pairwise IoU of $0.63$, corresponding to a median boundary difference of roughly two seconds on events averaging 23 seconds; the three answer groups agree on factual correctness for $79.8\%$ of questions; and $1.3\%$ of questions are rejected as invalid. To confirm that the procedure is stable, an author independently re-adjudicated 300 randomly sampled items, both accepted and rejected, and agreed with the production pipeline on $98\%$ of them.

\section{Prompts}
\label{App:prompt}

\begin{tcolorbox}[
    colback=gray!5,
    colframe=gray!50,
    title=Grounded VideoQA Prompt,
    fonttitle=\bfseries,
    breakable,
    enhanced,
    left=4pt, right=4pt, top=4pt, bottom=4pt,
]
{\small\ttfamily\raggedright%
After reviewing the video, provide the best answer to the following question.
Answer in 1-2 sentences.\\
Also provide the time interval (in seconds) where the answer evidence appears
in the video.\\[0.5em]

Use this exact format:\\
Answer: \textless your answer\textgreater\\
Time: [\textless start\_seconds\textgreater, \textless end\_seconds\textgreater]\\[0.5em]

\{question\}
}
\end{tcolorbox}

\begin{tcolorbox}[
    colback=gray!5,
    colframe=gray!50,
    title=Multi-Answer Judge Prompt,
    fonttitle=\bfseries,
    breakable,
    enhanced,
    left=4pt, right=4pt, top=4pt, bottom=4pt,
]
{\small\ttfamily\raggedright\obeylines%
You are an expert evaluator for video question answering systems.
Your task is to judge if a predicted answer is semantically correct compared to
ANY ONE of the provided ground-truth answers. The prediction is CORRECT if it
matches at least one ground-truth answer in meaning.
\vspace{0.5em}

Question: \{question\}
\vspace{0.5em}

Ground Truth Answers (prediction must match at least ONE):
\{gt\_formatted\}
\vspace{0.5em}

Predicted Answer: \{pred\_answer\}
\{type\_specific\_criteria\}
\vspace{0.5em}

Evaluation criteria:
1. The predicted answer should convey the same core meaning as at least ONE ground-truth answer.
2. Minor differences in phrasing, synonyms, or additional non-contradictory details are acceptable.
3. The core factual content must match at least one ground-truth answer.
4. Be lenient with minor variations but strict with factual errors or contradictions.
5. If the prediction says ``I cannot see the video'' or similar refusals, mark as INCORRECT.
\vspace{0.5em}

Hallucination detection:
6. Check if the prediction contains fabricated details NOT supported by any ground-truth answer.
\quad - Fabricated details include: invented objects, colors, locations, counts, actions, or events that do not appear in ANY of the ground-truth answers.
\quad - Generic hedging language (e.g., ``probably'', ``likely'', ``I think'') is NOT hallucination.
\quad - Reasonable elaborations that are logically implied by the ground-truth are NOT hallucination (e.g., GT says ``blue car'', prediction says ``a blue car parked on the street'' --- ``parked on the street'' is an added detail but not necessarily hallucination if plausible).
\quad - Specific invented details that contradict or go far beyond the ground-truth ARE hallucination (e.g., GT says ``blue car'', prediction says ``blue Ferrari with racing stripes'' --- ``Ferrari'' and ``racing stripes'' are fabricated specifics).
7. If the prediction contains the correct core answer BUT also includes hallucinated details:
\quad - If the hallucinated details are factually wrong or contradict the ground-truth, mark as INCORRECT. (e.g., GT says ``2 items'', prediction says ``2 items including a red apple and a banana'' when the actual items were different --- the count is right but the specifics are wrong, mark INCORRECT.)
\quad - If the hallucinated details are minor additions that don't contradict the ground-truth, mark as CORRECT but set has\_hallucination to true. (e.g., GT says ``blue car'', prediction says ``blue car on the left side'' --- ``on the left side'' is added but doesn't contradict anything.)
8. If the prediction is entirely hallucinated with no correct content, mark as INCORRECT.
\vspace{0.5em}

Respond in the following JSON format only (no markdown, no code blocks):
\{``is\_correct'': true/false, ``confidence'': 0.0-1.0, ``has\_hallucination'': true/false, ``matched\_answer\_index'': 1-based index of the matched GT answer or null if none matched, ``explanation'': ``brief explanation of your judgment, noting any hallucinated details if present''\}
\vspace{0.5em}

Your response:
}
\end{tcolorbox}

\end{document}